\documentclass[runningheads]{llncs}

\usepackage{eccv}

\usepackage{eccvabbrv}

\usepackage{graphicx}
\usepackage{booktabs}
\usepackage{svg} 
\usepackage{pifont}
\usepackage{multirow}
\usepackage[lined,boxed,commentsnumbered,ruled]{algorithm2e}

\usepackage[accsupp]{axessibility}  

\usepackage{hyperref}

\usepackage{orcidlink}

\begin{document}

\title{Unsupervised Variational Translator for Bridging Image Restoration and High-Level Vision Tasks} 

\titlerunning{Variational Translator}

\author{Jiawei Wu\inst{1}, \and
Zhi Jin\inst{1,2}*}

\authorrunning{J. Wu, Z. Jin.}

\institute{School of Intelligent Systems Engineering, Shenzhen Campus of Sun Yat-sen University, Shenzhen, Guangdong 518107, P.R.China\\ \and
Guangdong Provincial Key Laboratory of Fire Science and Technology, Guangzhou 510006, China\\
\email{wujw97@mail2.sysu.edu.cn}}

\maketitle
\renewcommand{\thefootnote}{}
\footnotetext[1]{* Corresponding author.}
\begin{abstract}
  Recent research tries to extend image restoration capabilities from human perception to machine perception, thereby enhancing the performance of high-level vision tasks in degraded environments. These methods, primarily based on supervised learning, typically involve the retraining of restoration networks or high-level vision networks. However, collecting paired data in real-world scenarios and retraining large-scale models are challenge. To this end, we propose an unsupervised learning method called \textbf{Va}riational \textbf{T}ranslator (VaT), which does not require retraining existing restoration and high-level vision networks. Instead, it establishes a lightweight network that serves as an intermediate bridge between them. By variational inference, VaT approximates the joint distribution of restoration output and high-level vision input, dividing the optimization objective into preserving content and maximizing marginal likelihood associated with high-level vision tasks. By cleverly leveraging self-training paradigms, VaT achieves the above optimization objective without requiring labels. As a result, the translated images maintain a close resemblance to their original content while also demonstrating exceptional performance on high-level vision tasks. Extensive experiments in dehazing and low-light enhancement for detection and classification show the superiority of our method over other state-of-the-art unsupervised counterparts, even significantly surpassing supervised methods in some complex real-world scenarios. Code is available at \url{https://github.com/Fire-friend/VaT}.
\end{abstract}

\section{Introduction}
\label{sec:intro}
\begin{figure*}[t]
    \centering
    \includegraphics[width=0.9\linewidth]{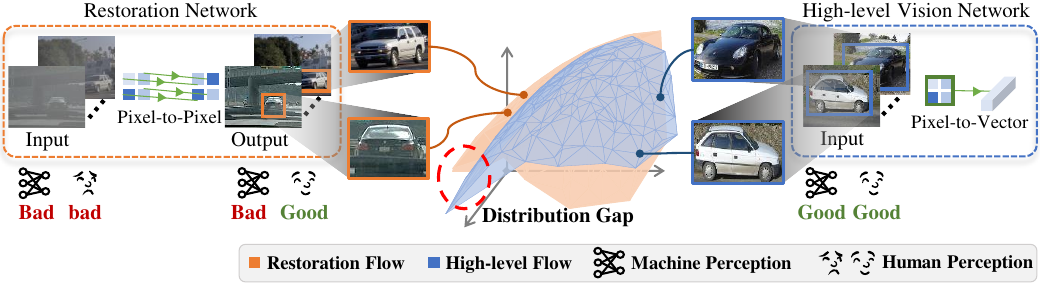}
    \caption{The fundamental insight that restoration cannot directly enhance high-level vision tasks. While restoration effectively enhances human perception by pixel-to-pixel, it may not improve machine perception. This discrepancy arises due to the distinct distribution gap between the restoration output and high-level vision input.}
    \label{fig:intro}
\end{figure*}
Image restoration aims to improve the quality of degraded images to enhance human perception \cite{banham1997digital}. Various sub-tasks of image restoration, such as dehazing \cite{berman2016non,yang2022self,song2023vision} and low-light image enhancement \cite{ma2022toward,wu2022uretinex,guo2023low}, have made considerable advancements. 
However, in applications involving high-level vision tasks (e.g., object detection in adverse weather), merely employing these restoration methods as pre-processing is proven inadequate \cite{pei2018does}.
This promotes the emergence of a significant research field, denoted as Task-Driven Image Restoration (TDIR), which aims to use restoration to enhance high-level vision tasks \cite{liu2022image,li2023detection,sun2022rethinking, yang2022self}.
 
Existing TDIR methods can be broadly categorized into union and separate categories \cite{li2023detection}. With high-level vision labels, union methods \cite{liu2018image,li2023detection,liu2020connecting, liu2022image} aim to co-train a hybrid network that integrates degradation networks as degradation modules, enabling them to effectively handle degraded images. These methods remove explicit supervision for image restoration to avoid conflicts \cite{li2023detection} with high-level vision tasks, resulting in a notable decrease in restoration performance as perceived by humans. In contrast, separate methods \cite{singh2019dual,dai2016image,li2017aod} tend to train restoration networks to produce clean images for pre-trained high-level vision models.
However, the restoration output may contain imperceptible noise that is harmful to subsequent high-level vision networks \cite{li2023detection}, often resulting in the inability to enhance the performance of high-level vision tasks \cite{liu2022image}. As shown in Figure \ref{fig:intro}, although restoration utilizes the same training set with the high-level vision tasks, a significant distribution gap persists between the restoration output and high-level vision input. 
To address this issue, Sun et al. \cite{sun2022rethinking} propose detection-friendly labels for fine-tuning restoration networks, yet models optimized for human perception only achieve suboptimal results.

Obviously, the performance of restoration and high-level vision tasks in existing methods exhibit a trade-off, primarily due to the misalignment between human perception and machine perception. Additionally, the aforementioned TDIR methods \cite{liu2018image,li2023detection,liu2020connecting, singh2019dual,dai2016image,li2017aod} presuppose that either the restoration or high-level vision models are ``trainable'', and paired training data, comprising degraded images, clean images, and high-level vision labels, are ``available''. However, these ``trainable'' and ``available'' aspects are often unattainable in numerous real-world scenarios. 
For example, the restoration model is a diffusion model \cite{ozdenizci2023restoring} and the high-level vision model is a large vision-language model \cite{radford2021learning}. Meanwhile, paired training data in real-world scenarios is often challenging \cite{liu2023degae}.
Therefore, the key challenge to advancing this field is \textit{how to efficiently bridge restoration and high-level vision tasks without requiring paired data and retraining existing networks.}

In this paper, our key insight is to model the true joint distribution of restoration output and high-level vision input. Using variational inference \cite{blei2017variational}, we derive an optimization objective with a content preservation term and a maximum likelihood term linked to high-level vision input. We found that our derived objective is identical to unpaired learning methods \cite{zhu2017unpaired}. This implies that existing unpaired learning methods \cite{park2020contrastive,xu2022maximum,jung2022exploring,zhu2017unpaired, xie2023unpaired} have the potential to serve as unsupervised methods that bridge restoration and high-level vision tasks. However, our experiments demonstrate that they struggle to deal with complex real-world degradation. Upon further analysis of the obtained bound using the full probability formula, we found that objects not relevant to high-level vision tasks hinder the maximum likelihood. Thus, we simplify the likelihood term to focus on relevant objects, obtaining a maximum marginal likelihood term. To achieve the new optimization objective in an unsupervised manner, we follow the standard unpaired learning \cite{zhu2017unpaired}, but replace the discriminator training with the proposed self-training strategy. Besides, instead of retraining the networks, we introduce a lightweight network to connect them, with a gated fusion module to address restoration failures and a transformation module for image translation.

Extensive experiments demonstrate that VaT can significantly enhance the performance of high-level vision tasks without the retraining of existing restoration and high-level vision models. In the task of object detection under foggy, VaT outperforms unsupervised counterparts \cite{park2020contrastive,xu2022maximum,jung2022exploring,zhu2017unpaired, xie2023unpaired} by 10\% of mean Average Precision (mAP), and even surpass supervised methods by around 4\% of mAP on real degraded. Our contributions include: (i) We derive an upper bound for modeling the joint distribution of restoration output and high-level vision input and further simplify it to focus on the interested objects of high-level vision tasks. (ii) We propose VaT, the first unsupervised method for bridging restoration and high-level vision tasks, effectively achieving our optimization objective through uncertainty-guided self-training. (iii) For the first time, we connect the restoration and high-level vision networks without retraining by inserting the proposed lightweight network. It includes a gated fusion module to tackle restoration failures and a transformation module for image translation.

\section{Related work}
\subsection{Restoration for High-level Vision Tasks}
In degraded environments, it is established that high-level vision tasks can achieve superior performance on enhanced images through image restoration. In the early stage, Li et al. \cite{li2017aod} use the proposed AODNet as preprocessing to successfully enhance the performance of object detection under synthetic foggy conditions. However, Pei et al. \cite{pei2018does} argue that dehazing methods as preprocessing struggle to improve the performance of high-level vision tasks in real-world scenarios and may even exacerbate the issue. Therefore, researchers begin to focus on integrating restoration as modules to improve the performance of high-level vision networks in degraded environments \cite{li2023detection, liu2022image, yang2022self}. Due to the absence of restoration supervision, these methods significantly diminish the restoration quality for human perception. Sun et al. \cite{sun2022rethinking} use adversarial attacks to embed imperceptible information preferences for detection into restoration labels, enhancing high-level vision tasks while maintaining image restoration performance. However, their experiments show that the restoration network has difficulty fitting to these labels, resulting in only a slight improvement in high-level vision performance. Therefore, current research cannot effectively connect image restoration with high-level vision tasks \cite{sun2022rethinking, liu2022image, yang2022self, li2023detection}. Different from them, our goal is not to retrain both the restoration and high-level vision models, but rather to employ a lightweight network that connects pre-trained ones.

\subsection{Unparied Image-to-image Translation}
With unpaired data, there can be infinite ways to map two domains. To address this issue, many effective solution paradigms have been proposed. Relationship preservation methods \cite{benaim2017one,fu2019geometry,han2021dual,xie2021unaligned,xie2022unsupervised} encourage relationships present in the input to be analogously reflected in the output. Shared latent space methods \cite{huang2018multimodal,lee2018diverse,liu2018unified} employ the assumption that there exists a shared latent code in the latent space for any given paired images. Contrastive learning methods \cite{park2020contrastive,han2021dual,jia2021semantically,jung2022exploring} assume that the two corresponding patches in the input and output images should have larger mutual information than others. Notably, the most well-known paradigm is cycle-consistency methods \cite{chen2022vector,dalva2022vecgan,tang2021attentiongan,zhu2017unpaired}, which enforces the network to be a one-to-one mapping by cycle-consistency training. In this paper, we also adopt the paradigm of cycle consistency to leverage unpaired data. However, unlike directly restricting input to output, we constrain the cycle consistency between intermediate results and network output in our approach.

\section{Variational Perspective}

\subsection{Preliminaries}
Given unpaired degraded images $I_{LQ}$ and high-quality images $I_{HQ}$, a restoration network $\mathcal R$ trained on unknown degraded images, and a high-level vision network $\mathcal D$ trained on $I_{HQ}$ with corresponding labels, our goal is to build the joint distribution $P(I_R, I_{HQ})$, where $I_R=\mathcal R(I_{LQ})$ denotes the restoration output. Once the joint distribution is obtained, the high-level vision model-friendly input can be easily obtained from the restoration output. We further found that establishing the joint distribution has an important advantage.
\begin{remark}
    \label{remark:1}
    (\textit{The better the restoration, the better the effect on high-level vision tasks.}) \rm 
    Consider the two primary objectives of restoration: achieving high generalization and high performance \cite{liu2023evaluating}. For generalization, if the restoration model demonstrates superior generalization performance, implying that the marginal distribution $P(I_R)$ closely approximates the true distribution. Since $P(I_{HQ})$ is known, this results in the joint distribution $P(I_R, I_{HQ})$ being closer to the true distribution, i.e., having improved generalization, too. As for performance, a higher-performing restoration model can mitigate the challenge of establishing the joint distribution, thereby enhancing its effectiveness.
\end{remark}
Accordingly, the joint distribution $P(I_R, I_{HQ})$ serves as an excellent bridge to make restoration promote high-level vision tasks. However, we often struggle to obtain the input of high-level vision tasks corresponding to the image restoration output, obtaining the analytic expression of $P(I_R, I_{HQ})$ in practice is unfeasible.

\subsection{Variational Objective of Joint Distribution}
To solve the joint distribution $P(I_R, I_{HQ})$, we approximate it using a parameterized distribution $Q_\theta(I_R, I_{HQ})$. The goal is to minimize their Kullback–Leibler divergence \cite{kullback1951information} as follows:
\begin{equation}
\begin{split}
\mathcal L &= \min KL[Q_{\theta}(I_R, I_{HQ})||P(I_R, I_{HQ})]\\
&=\iint Q_{\theta}(I_R, I_{HQ})\log \frac{Q_\theta(I_R, I_{HQ})}{P(I_R, I_{HQ})} dI_{HQ} d I_R.
\end{split}
\end{equation}
According to variational inference \cite{blei2017variational} which is detailed in Appendix, we can obtain the final objective as
\begin{equation}
\begin{split}
\mathcal L=\min\mathbb E_{I_R \sim Q_\theta(I_R)}&[\mathbb E_{I_{HQ} \sim Q_\theta(I_{HQ}|I_R)}( \\ &\underbrace{-\log P(I_R|I_{HQ})}_{\text{\ding{172} Reconstruction}} \underbrace{-\log P(I_{HQ})}_{\substack{\text{\ding{173} Maximum} \\ \text{likelihood}}}\underbrace{+\mathcal H(Q_\theta(I_{HQ}|I_R)}_{\substack{\text{\ding{174} Entropy} \\ \text{minimization}}})],
\end{split}
\label{eq:objective1}
\end{equation}
where $Q_\theta(I_{HQ} | I_R)$ can be modeled by a image-to image translation network as described in Section \ref{section:VaT}. Then, the Eq. \ref{eq:objective1} can be further analyzed as follows:

\noindent\ding{172} \textbf{Reconstruction.} The first term demonstrates that the translated images from $Q_\theta(I_{HQ} | I_R)$ needs to satisfy the conditional distribution $P(I_R | I_{HQ})$. It is a regularization technique referred to as reconstruction loss \cite{kingma2014auto}, ensuring the invariant contents during the image-to-image translation. 

\noindent\ding{173} \textbf{Maximum likelihood.} The second term expects translated images to maximize the likelihood of the distribution $P(I_{HQ})$ of high-level vision input. Thus it's the key to promoting restoration output adapt to the high-level vision input. 

\noindent\ding{174} \textbf{Entropy minimization.} The third term expects the network can make low-entropy predictions for keeping confident. Most of the time, it can be ignored since it only boosts confidence without impacting accuracy.

\subsection{Analysis of Variational Objective}
Based on the variational objective of Eq. \ref{eq:objective1}, we can analyze the relationship with unpaired learning methods \cite{zhu2017unpaired, park2020contrastive}, which are potential unsupervised methods to bridge the restoration and high-level vision tasks. Surprisingly, the objective of Eq. \ref{eq:objective1} is highly similar to them. The cycle-consistency loss \cite{zhu2017unpaired} and the patchwise contrastive loss \cite{park2020contrastive} satisfy the first term, and the adversarial discriminator satisfies the second term. These unpaired methods already achieve remarkable performance in many tasks \cite{chen2022unpaired, chen2022unpaired1, pang2021image}, further verifying the effectiveness of building joint distributions. However, our experiments in Section \ref{ex:results} demonstrate that these methods often fail for the TDIR task in complex real-world degradation. The reason can be concluded from the maximum likelihood term of Eq. \ref{eq:objective1}, which can be expanded using the total probability formula as
\begin{equation}
\begin{split}
\underbrace{-\log P(I_{HQ})}_{\text{\ding{173} Maximum likelihood}} = -\log \sum _{i=1}^n P(I_{HQ}|y_d^i) P(y_d^i), s.t. I_{HQ} \sim Q_\theta(I_{HQ}|I_R),
\end{split}
\label{eq:objective}
\end{equation}
where $n$ denotes the number of instance objects in the whole dataset, and $y_d$ denotes the interested objects of high-level vision tasks. For a high-level vision task that interests $m$ objects, maximum likelihood requires unpaired learning methods to accurately identify and map the $m$ interested objects from $n$ objects. However, for the TDIR task in complex real-world scenarios, the number of interested objects is much smaller than the number of all objects, i.e., $m << n$, resulting in unpaired learning methods hard to complete accurate mapping among many interfering objects. Therefore, the original objective of unpaired learning methods (Eq. \ref{eq:objective}) only considers the global mapping of two domains. 

To focus on the interested objects of high-level vision tasks, we neglect the objects of interfering objects to maximize the marginal likelihood as follows:
\begin{equation}
\begin{split}
\mathcal L=\min\mathbb E_{I_R \sim Q_\theta(I_R)}[\mathbb E_{I_{HQ} \sim Q_\theta(I_{HQ}|I_R)}(\underbrace{-\log P(I_R|I_{HQ})}_{\text{\ding{172} Reconstruction}}&\underbrace{-\log \sum_{i=1}^mP(I_{HQ}|y_d^i)}_{\substack{\text{\ding{173} Maximum marginal} \\ \text{likelihood}}})],
\label{eq:objective_m}
\end{split}
\end{equation}
where $m$ denotes the number of interested objects for high-level vision tasks. The complete derivation process can be found in the Appendix. To further verify the effectiveness of Eq \ref{eq:objective_m}, we discuss its relationship with existing TDIR methods.
\begin{remark}
    \label{remark:2}
    (\textit{The condition for translated images to satisfy both human and machine perception is to satisfy Eq. \ref{eq:objective_m}.}) By variable substitution, i.e.,$I_R \rightarrow I_{LQ}$ and $I_{HQ} \rightarrow Y$, Eq. \ref{eq:objective_m} equals to $\min\mathbb E_{Y \sim Q_\theta(Y|I_{LQ})}(-\log P(I_{LQ}|Y)-\log \sum_{i=1}^mP(Y))$. Obviously, the first term (reconstruction) is the optimization objective of existing separate TDIR methods \cite{dai2016image,li2017aod,singh2019dual}, while the second term (maximum marginal likelihood) is the optimization objective of existing union TDIR methods \cite{liu2018image,li2023detection,liu2020connecting, liu2022image}. Therefore, we can conclude that the reason union methods struggle to generate human perception-friendly images is the absence of the first term, whereas the absence of the second term hinders separate methods from improving high-level vision performance. In a novel approach, based on the mind of separate methods, Sun et al. \cite{sun2022rethinking} introduce the adversarial attack to explicitly generate human-machine-friendly images as restoration labels for retraining. This method effectively embodies the aforementioned equation. Thus it can take into account the performance of both high-level vision and restoration tasks. Therefore, the key to bridging high-level vision and restoration is how to realize the Eq. \ref{eq:objective_m}. 
\end{remark}

\section{Unsupervised Variational Translator}
\label{section:VaT}
\begin{figure*}[t]
    \centering
    \includegraphics[width=0.9\linewidth]{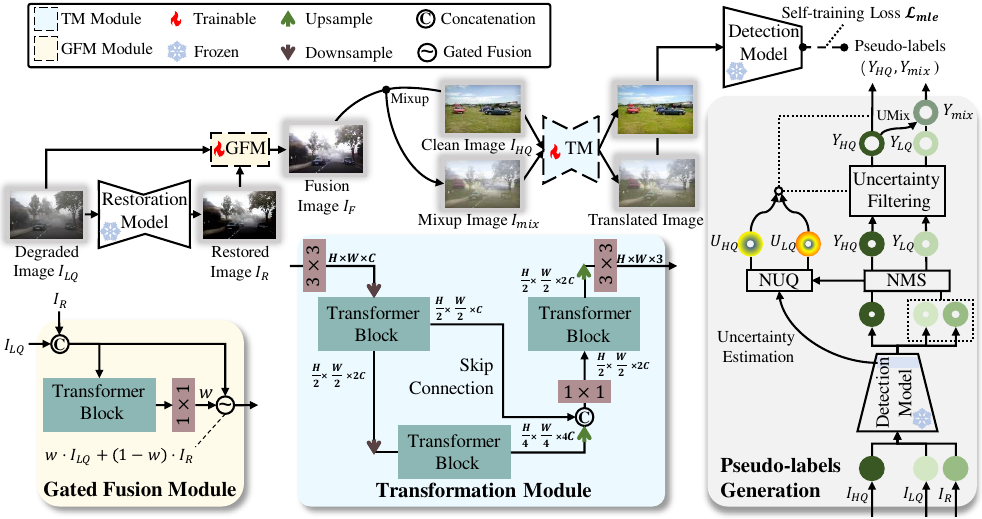}
    \caption{Overview of our variational translator (VaT). Uncertainty-guided pseudo-label generation is used to obtain pseudo-labels for degraded images. The Gated Fusion Module (GFM) combines degraded and restored images, which are then fed into the Transformation Module (TM). The acquired pseudo-labels are then employed to supervise the detection prediction of translated images augmented by mixup.}
    \label{fig:method}
\end{figure*}

In this section, we introduce the proposed unsupervised Variational Translator (VaT) as shown in Figure \ref{fig:method}, aimed at efficiently achieving the objective in Eq. \ref{eq:objective_m}. We first introduce the network architecture of our VaT and then present the unsupervised optimization method towards the objective.

\subsection{Architecture of VaT}
Intuitively, one might consider using the restored images as the input for VaT. However, due to the limited generalization performance of the restoration network, its restoration output may be unsatisfactory in practice, and in some cases, the restoration output even deteriorates compared to the original degraded images \cite{pei2018does}. In such scenarios, as per \textit{Remark \ref{remark:1}}, initiating the transformation from the degraded input rather than the inferior restoration output could yield better outcomes. To address this issue, we propose a gated fusion module, which adaptively combines the degraded input and restoration output by the network itself, to obtain optimal initial images for transformation. The degraded images and restoration images are fed into a transformer block layer \cite{vaswani2017attention} and a point convolutional layer with one output channel to derive weights $w$. It undergoes processing via a sigmoid activation function $\sigma(\cdot)$ to serve as gated fusion weights, efficiently generating the fusion image $I_F$ as follows:
\begin{equation}
    I_F = \sigma(w) \odot I_{LQ} + (1- \sigma(w)) \odot I_R.
\end{equation}

Subsequently, we introduce a straightforward U-shape transformation module, leveraging the transformer block \cite{vaswani2017attention}, to adeptly convert these images into high-level vision model-friendly images. This U-shape architecture preserves image details via skip connections while simultaneously altering semantics through the encoder-decoder framework \cite{ronneberger2015u}. The transformer block augments learning capabilities by furnishing a global receptive field, empowering neural networks to grasp context information \cite{ronneberger2015u}. Although both modules are straightforward, they possess the characteristics necessary for bridging restoration and high-level vision tasks. The two module details can be shown in Figure \ref{fig:method}.




\begin{figure}[t]
    \centering
    \includegraphics[width=0.9\linewidth]{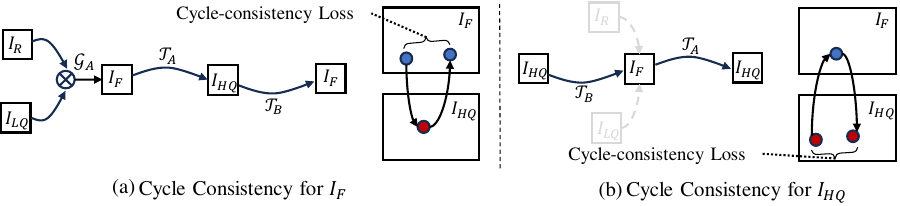}
    \caption{Our model tends to contain the cycle-consistency between fusion output $I_F$ and VaT output $I_{HQ}$, including three mapping functions $\mathcal G_A: I_R, I_{LQ} \rightarrow I_F$, $\mathcal T_A:I_F \rightarrow I_{HQ}$, and $\mathcal T_B:I_{HQ} \rightarrow I_F$. (a) Forward cycle-consistency loss: $\mathcal R_A(I_R, I_{LQ}) \rightarrow \mathcal T_A(\mathcal G_A(I_R, I_{LQ})) \rightarrow \mathcal T_B(\mathcal T_A(\mathcal G_A(I_R, I_{LQ}))) \approx \mathcal G_A(I_R, I_{LQ})$. (b) Backward cycle-consistency loss: $I_{HQ} \rightarrow \mathcal T_B(I_{HQ}) \rightarrow \mathcal T_A(\mathcal T_B(I_{HQ})) \approx I_{HQ}$.
    \label{fig:cycle}}
\end{figure}
\subsection{VaT Optimization}
Then, by Remark \ref{remark:2}, we focus on the VaT optimization to satisfy Eq. \ref{eq:objective_m} as follows:

\noindent \ding{172} \textbf{Reconstruction.}
Firstly, we focus on how to satisfy the first term of Eq. \ref{eq:objective_m}. There are many effective manners \cite{zhu2017unpaired, park2020contrastive, xie2023unpaired} in unpaired learning methods that can achieve it, in which cycle-consistency loss \cite{zhu2017unpaired} is the most commonly used one. Accordingly, we introduce it into VaT. Due to VaT containing the gated fusion module, the original cycle consistency training cannot directly be applied to it. To address this issue, we modify the cycle consistency training process for VaT. Specifically, our model tends to contain the cycle-consistency between the fusion output $I_F$ and the VaT output $I_{HQ} \sim Q_{\theta}(I_{HQ}|I_{R})$, including three mapping functions $\mathcal G_A: I_R, I_{LQ} \rightarrow I_F$, $\mathcal T_A:I_F \rightarrow I_{HQ}$, and $\mathcal T_B:I_{HQ} \rightarrow I_F$. Following the original cycle-consistency loss \cite{zhu2017unpaired}, we introduce two cycle consistency loss terms to further regularize the mapping as follows:
\begin{equation}
    \mathcal L_{cyc}(I_{LQ}, I_R) = \parallel I_F - \mathcal T_B(\mathcal T_A(I_F))\parallel_1 + \parallel I_{HQ} - \mathcal T_A(\mathcal T_B(I_{HQ}))\parallel_1,
    \label{eq:cycle}
\end{equation}
where $I_F = \mathcal G_A(I_R, I_{LQ})$, $\mathcal G_A$ is the gated fusion module, $\mathcal T_A, \mathcal T_B$ are two transformation modules with the same structure. Therefore, the full VaT model is employed from A to B, whereas the VaT model without the gated fusion module is applied from B to A as shown in Figure \ref{fig:cycle}.

\noindent \ding{173} \textbf{Maximum Marginal likelihood.} Then, we focus on how to satisfy the second term of Eq. \ref{eq:objective_m} by unsupervised learning. Since minimizing the predictive risk is an effective way to maximize likelihood \cite{yosida2012functional}, improving the performance of high-level vision outputs can achieve maximum marginal likelihood. 

\begin{algorithm}[tb]
\resizebox{0.9\linewidth}{!}{
\begin{minipage}{\linewidth}
\begin{small}
\caption{\small Training process of VaT.}
\label{alg:Framwork} 
\SetAlgoLined
\KwIn{Degraded training set $\{I_{LQ}^i\}_{i=1}^N$, clean training set $\{I_{HQ}^i\}_{j=1}^M$, pre-trained restoration model $\mathcal R$, pre-trained high-level vision model $\mathcal D$.}
\textbf{Initialization:} Initialize the parameters $\theta_A$, $\theta_B$ of $\mathcal F_{VaT}^A$ ,$\mathcal F_{VaT}^B$, where $\mathcal F_{VaT}^*$ is composed of a gated fusion module $\mathcal G_* $ and a transformation module $\mathcal T_*$. \\ 
    \For {each iteration}{
        Obtain restoration prediction \small $I_R^i=\mathcal R(I_{LQ}^i)$.\\
        Calculate cycle loss \small $\mathcal L_{cyc}(I_{LQ}^i,I_R^i)$ by Eq. \ref{eq:cycle}.\\
        Obtain paired labels of clean training set \small{$Y_{HQ}=\mathcal D(I_{HQ}^i)$.} \\
        Obtain pseudo-labels of degraded training set $Y_{LQ}= \text{NMS}(\mathcal D(I_{LQ}^i), \mathcal D(I_R^i))$ of degraded image by the principle of maximum predicted confidence.\\
        Obtain risk $U_{LQ}$ of pseudo-labels $Y_{LQ}$ by NUQ \cite{kotelevskii2022nonparametric}.\\
        Obtain filtered pseudo-labels $Y_{LQ} = \mathbb I_{U_{LQ} < \epsilon} Y_{LQ}$. \\
        Generate strong augmented samples $I_{mix}, Y_{mix}$ using mixup by Eq.\ref{eq:Umix}.\\
        Calculate high-level vision loss $\mathcal L_{h}$ by Eq. \ref{eq:high-level}.\\
        Update model parameters $\theta_A,\theta_B$ to minimize loss $\mathcal L = \mathcal L_{cyc}+\mathcal L_{h}$.
    }
\KwOut{VaT model $\mathcal F_{VaT}^A(\theta_A)$.}
\end{small}
\end{minipage}
}
\end{algorithm}

Inspired by the self-training paradigm \cite{sohn2020fixmatch}, we propose an effective unsupervised optimization pipeline to improve the high-level vision outputs. For clean images $I_{HQ} \sim P(I_{HQ})$, it belongs to the same domain as the training data of the high-level vision model, thus we utilize the predictions of the high-level vision model as the high-level vision task labels $Y_{HQ}$. For degraded images $I_{LQ}$, we can obtain its corresponding enhanced results $I_R={\mathcal R}(I_{LQ})$. The predictions of the high-level vision model for $I_{LQ}$ and $I_R$ may all contain potential true labels. To fully exploit these potential true labels, we use non-parametric estimation of uncertainty \cite{kotelevskii2022nonparametric} to filter these predictions and consider low-uncertainty predictions as pseudo-labels $Y_{LQ}$ for degraded images $I_{LQ}$. Then, drawing inspiration from the idea that the prediction of weak augmentation can supervise strong ones \cite{sohn2020fixmatch}, we construct strong augmented data using mixup \cite{zhang2018mixup, han2022umix} guided by uncertainty as
\begin{equation}
\begin{split}
    I_{mix}&=\lambda \cdot \mathcal G(I_{LQ},I_R) + (1-\lambda) \cdot I_{HQ},\\
    Y_{mix}&= U_{LQ} \cdot \lambda \cdot Y_{LQ} + U_{HQ}\cdot (1-\lambda) \cdot Y_{HQ},
    \label{eq:Umix}
\end{split}
\end{equation}
where $\lambda \sim Beta(\alpha, \alpha)$, $\mathcal G$ denotes the gated fusion module, $U_{LQ}$ and $U_{HQ}$ represent the predicted uncertainty of the high-level vision predictions for degraded and clean images respectively, $Y_{LQ}$ and $Y_{HQ}$ denote the pseudo-labels assigned to $I_{LQ}$ and $I_{HQ}$. Then, we use the original high-level vision loss $\mathcal L_h$ to constrain high-level vision model predictions as follows:
\begin{equation}
    \mathcal L_{mle}=\mathcal L_h(\mathcal F_{VaT}(I_{HQ}), Y_{HQ}) + \mathcal L_h(\mathcal F_{VaT}(I_{mix}), Y_{mix}).
    \label{eq:high-level}
\end{equation}
where $\mathcal F_{VaT}$ denotes the proposed network. For high-level vision tasks, which may involve dense predictions (e.g., object detection), pseudo-labels could miss potential true labels. Therefore, for dense tasks, we only calculate the second loss term for labels that contain the target. The training process of VaT can be seen in Algorithm \ref{alg:Framwork}. More details of training can be found in the Appendix.


 

\begin{table*}[tb]
\centering
\caption{Quantitative comparison results of RTTS and VOCh-test, $\ddagger$ adopts publicly available AODNet pre-trained weights, while $\dagger$ adopts weights retrained on VOCh set. \textcolor{red}{Red} denotes the best results, while \textcolor{blue}{blue} denotes the second-best results.}
\label{tab:COM}
\resizebox{\linewidth}{!}{
\begin{tabular}{|c|l|cc|cc|cc|cc|cc|cc}
\hline
\multirow{2}{*}{}& Dataset & \multicolumn{6}{c|}{RTTS} & \multicolumn{6}{c|}{VOCh-test} \\ \cline{2-14} 
& Class & \multicolumn{1}{c}{bicycle} & \multicolumn{1}{c}{bus} & \multicolumn{1}{c}{car} & \multicolumn{1}{c}{motor}  & \multicolumn{1}{c|}{person} & \multicolumn{1}{c|}{mAP $\uparrow$} & \multicolumn{1}{c}{bicycle} & \multicolumn{1}{c}{bus} & \multicolumn{1}{c}{car} & \multicolumn{1}{c}{motor}  & \multicolumn{1}{c|}{person} & \multicolumn{1}{c|}{mAP $\uparrow$} \\ \hline
\multirow{6}{*}{\rotatebox{90}{Supervised}} & Vanilla \cite{redmon2018yolov3} & \multicolumn{1}{c}{43.57}  &  \multicolumn{1}{c}{16.09} &  \multicolumn{1}{c}{50.59} & \multicolumn{1}{c}{40.18} & \multicolumn{1}{c|}{66.81} &\multicolumn{1}{c|}{43.45} & \multicolumn{1}{c}{81.66} & \multicolumn{1}{c}{81.24}  & \multicolumn{1}{c}{84.29}  & \multicolumn{1}{c}{80.10}  & \multicolumn{1}{c|}{76.16} &  \multicolumn{1}{c|}{80.69} \\ 
& Baseline $\ddagger$ & \multicolumn{1}{c}{29.64} & \multicolumn{1}{c}{11.96} &  \multicolumn{1}{c}{43.67} & \multicolumn{1}{c}{27.35} &  \multicolumn{1}{c|}{60.08} & \multicolumn{1}{c|}{34.54} & \multicolumn{1}{c}{79.47} & \multicolumn{1}{c}{74.14} & \multicolumn{1}{c}{81.73} & \multicolumn{1}{c}{77.81} & \multicolumn{1}{c|}{72.25} & \multicolumn{1}{c|}{77.08} \\ 
& Baseline $\dagger$  & \multicolumn{1}{c}{34.31} & \multicolumn{1}{c}{12.71} &  \multicolumn{1}{c}{45.14} & \multicolumn{1}{c}{31.83} &  \multicolumn{1}{c|}{62.50} & \multicolumn{1}{c|}{37.30} & \multicolumn{1}{c}{81.21} & \multicolumn{1}{c}{77.42} & \multicolumn{1}{c}{84.51} & \multicolumn{1}{c}{80.39} & \multicolumn{1}{c|}{76.64} & \multicolumn{1}{c|}{80.04} \\
& CT-YOLO & \multicolumn{1}{c}{\textcolor{blue}{45.30}} & \multicolumn{1}{c}{14.62} &  \multicolumn{1}{c}{52.89} & \multicolumn{1}{c}{41.37} &  \multicolumn{1}{c|}{67.74} & \multicolumn{1}{c|}{44.38} & \multicolumn{1}{c}{84.43} & \multicolumn{1}{c}{77.68} & \multicolumn{1}{c}{86.27} & \multicolumn{1}{c}{82.39} & \multicolumn{1}{c|}{79.46} & \multicolumn{1}{c|}{82.05} \\
& IA-YOLO \cite{liu2022image} & \multicolumn{1}{c}{42.89} & \multicolumn{1}{c}{14.21} &  \multicolumn{1}{c}{51.62} & \multicolumn{1}{c}{37.68} &  \multicolumn{1}{c|}{66.72} & \multicolumn{1}{c|}{42.62} & \multicolumn{1}{c}{\textcolor{blue}{85.11}} & \multicolumn{1}{c}{81.34} &  \multicolumn{1}{c}{86.09} & \multicolumn{1}{c}{\textcolor{red}{84.66}} &  \multicolumn{1}{c|}{80.29} & \multicolumn{1}{c|}{83.50} \\ 
& AD-YOLO \cite{sun2022rethinking}& \multicolumn{1}{c}{\textcolor{red}{46.21}} & \multicolumn{1}{c}{14.56} &  \multicolumn{1}{c}{54.09} & \multicolumn{1}{c}{42.33} &  \multicolumn{1}{c|}{70.16} & \multicolumn{1}{c|}{45.47} & \multicolumn{1}{c}{\textcolor{red}{85.45}} & \multicolumn{1}{c}{83.21} & \multicolumn{1}{c}{87.01} & \multicolumn{1}{c}{\textcolor{blue}{84.45}} & \multicolumn{1}{c|}{80.42} & \multicolumn{1}{c|}{\textcolor{red}{84.11}} \\ \hline  
\multirow{12}{*}{\rotatebox{90}{Unsupervised}} & CycleGAN $\ddagger$ \cite{zhu2017unpaired}& \multicolumn{1}{c}{34.46} &  \multicolumn{1}{c}{6.73} &  \multicolumn{1}{c}{31.76} & \multicolumn{1}{c}{20.86} & \multicolumn{1}{c|}{53.62} & \multicolumn{1}{c|}{29.29} & \multicolumn{1}{c}{82.04} &\multicolumn{1}{c}{80.34}  & \multicolumn{1}{c}{85.61} & \multicolumn{1}{c}{82.16} & \multicolumn{1}{c|}{\textcolor{blue}{82.16}} & \multicolumn{1}{c|}{81.60} \\ 
& CUT $\ddagger$ \cite{park2020contrastive}& \multicolumn{1}{c}{27.25} & \multicolumn{1}{c}{12.34} &  \multicolumn{1}{c}{34.51} & \multicolumn{1}{c}{24.28} &  \multicolumn{1}{c|}{55.39} & \multicolumn{1}{c|}{30.35} & \multicolumn{1}{c}{84.25} & \multicolumn{1}{c}{82.45} & \multicolumn{1}{c}{83.82} & \multicolumn{1}{c}{81.11} & \multicolumn{1}{c|}{81.84} & \multicolumn{1}{c|}{82.69} \\ 
& MSPC $\ddagger$ \cite{xu2022maximum}& \multicolumn{1}{c}{31.45} &  \multicolumn{1}{c}{9.13} &  \multicolumn{1}{c}{37.97} & \multicolumn{1}{c}{23.05} & \multicolumn{1}{c|}{52.34} & \multicolumn{1}{c|}{30.78} & \multicolumn{1}{c}{82.11} &\multicolumn{1}{c}{\textcolor{blue}{84.38}}  & \multicolumn{1}{c}{80.41} & \multicolumn{1}{c}{80.37} & \multicolumn{1}{c|}{81.23} & \multicolumn{1}{c|}{81.70} \\
& SRC $\ddagger$ \cite{jung2022exploring}& \multicolumn{1}{c}{35.69} &  \multicolumn{1}{c}{11.35} &  \multicolumn{1}{c}{37.86} & \multicolumn{1}{c}{23.75} & \multicolumn{1}{c|}{55.48} & \multicolumn{1}{c|}{32.82} & \multicolumn{1}{c}{82.86} &\multicolumn{1}{c}{82.41}  & \multicolumn{1}{c}{85.93} & \multicolumn{1}{c}{81.88} & \multicolumn{1}{c|}{80.96} & \multicolumn{1}{c|}{82.81} \\
& SPR $\ddagger$ \cite{xie2023unpaired}& \multicolumn{1}{c}{35.91} &  \multicolumn{1}{c}{12.76} &  \multicolumn{1}{c}{36.39} & \multicolumn{1}{c}{25.74} & \multicolumn{1}{c|}{56.77} & \multicolumn{1}{c|}{33.51} & \multicolumn{1}{c}{82.27} &\multicolumn{1}{c}{{83.49}}  & \multicolumn{1}{c}{87.09} & \multicolumn{1}{c}{81.86} & \multicolumn{1}{c|}{81.37} & \multicolumn{1}{c|}{83.21} \\
& VaT (Ours) $\ddagger$ & \multicolumn{1}{c}{43.21} & \multicolumn{1}{c}{\textcolor{blue}{24.55}} &  \multicolumn{1}{c}{\textcolor{red}{63.69}} & \multicolumn{1}{c}{\textcolor{blue}{43.77}} &  \multicolumn{1}{c|}{\textcolor{red}{72.19}} & \multicolumn{1}{c|}{\textcolor{blue}{49.48}} & \multicolumn{1}{c}{82.41} & \multicolumn{1}{c}{82.38} & \multicolumn{1}{c}{\textcolor{blue}{88.18}} & \multicolumn{1}{c}{83.01} & \multicolumn{1}{c|}{81.96} & \multicolumn{1}{c|}{83.59} \\ \cline{2-14}

& CycleGAN $\dagger$ \cite{zhu2017unpaired}& \multicolumn{1}{c}{33.12} & \multicolumn{1}{c}{8.49} &  \multicolumn{1}{c}{36.71} & \multicolumn{1}{c}{21.74} &  \multicolumn{1}{c|}{54.99} & \multicolumn{1}{c|}{31.01} & \multicolumn{1}{c}{83.01} & \multicolumn{1}{c}{79.25} & \multicolumn{1}{c}{85.50} & \multicolumn{1}{c}{80.42} & \multicolumn{1}{c|}{77.32} & \multicolumn{1}{c|}{81.10} \\ 
& CUT $\dagger$ \cite{park2020contrastive}& \multicolumn{1}{c}{29.84} & \multicolumn{1}{c}{9.77} &  \multicolumn{1}{c}{40.40} & \multicolumn{1}{c}{23.66} &  \multicolumn{1}{c|}{55.90} & \multicolumn{1}{c|}{31.92} & \multicolumn{1}{c}{72.19} & \multicolumn{1}{c}{73.54} & \multicolumn{1}{c}{80.89} & \multicolumn{1}{c}{73.81} & \multicolumn{1}{c|}{73.21} & \multicolumn{1}{c|}{74.73} \\
& MSPC $\dagger$ \cite{xu2022maximum}& \multicolumn{1}{c}{27.15} &  \multicolumn{1}{c}{9.83} &  \multicolumn{1}{c}{39.02} & \multicolumn{1}{c}{24.12} & \multicolumn{1}{c|}{55.64} & \multicolumn{1}{c|}{31.15} & \multicolumn{1}{c}{81.09} &\multicolumn{1}{c}{76.70}  & \multicolumn{1}{c}{84.71} & \multicolumn{1}{c}{80.41} & \multicolumn{1}{c|}{77.35} & \multicolumn{1}{c|}{80.05} \\
& SRC $\dagger$ \cite{jung2022exploring}& \multicolumn{1}{c}{34.28} &  \multicolumn{1}{c}{10.12} &  \multicolumn{1}{c}{38.96} & \multicolumn{1}{c}{28.16} & \multicolumn{1}{c|}{58.33} & \multicolumn{1}{c|}{33.97} & \multicolumn{1}{c}{77.89} &\multicolumn{1}{c}{79.02}  & \multicolumn{1}{c}{84.12} & \multicolumn{1}{c}{81.60} & \multicolumn{1}{c|}{79.17} & \multicolumn{1}{c|}{80.36} \\
& SPR $\dagger$ \cite{xie2023unpaired}& \multicolumn{1}{c}{29.57} &  \multicolumn{1}{c}{16.81} &  \multicolumn{1}{c}{45.28} & \multicolumn{1}{c}{27.93} & \multicolumn{1}{c|}{59.79} & \multicolumn{1}{c|}{35.87} & \multicolumn{1}{c}{81.38} &\multicolumn{1}{c}{81.17}  & \multicolumn{1}{c}{85.95} & \multicolumn{1}{c}{81.69} & \multicolumn{1}{c|}{79.19} & \multicolumn{1}{c|}{81.87} \\
& VaT (Ours) $\dagger$ & \multicolumn{1}{c}{42.69} & \multicolumn{1}{c}{\textcolor{red}{24.64}} &  \multicolumn{1}{c}{\textcolor{blue}{63.63}} & \multicolumn{1}{c}{\textcolor{red}{45.41}} &  \multicolumn{1}{c|}{\textcolor{blue}{71.97}} & \multicolumn{1}{c|}{\textcolor{red}{49.67}} & \multicolumn{1}{c}{81.98} & \multicolumn{1}{c}{\textcolor{red}{84.40}} & \multicolumn{1}{c}{\textcolor{red}{88.32}} & \multicolumn{1}{c}{83.23} & \multicolumn{1}{c|}{\textcolor{red}{82.18}} & \multicolumn{1}{c|}{\textcolor{blue}{84.02}} \\ \hline 
\end{tabular}
}
\end{table*}

\begin{table}[tb]
\centering
\caption{Quantitative comparison results of visual quality on RTTS and VOCh-test. \textcolor{red}{Red} denotes the best results, while \textcolor{blue}{blue} denotes the second-best results.}
\label{tb:visual}
\resizebox{0.95\linewidth}{!}{
\begin{tabular}{@{}l|r|c|c|c|c|c|c|c@{}}
\toprule
     &       & Vanilla \cite{li2017aod} & AD-YOLO \cite{sun2022rethinking}& CycleGAN \cite{zhu2017unpaired} & MSPC \cite{xu2022maximum} & SRC \cite{jung2022exploring} & SPR \cite{xie2023unpaired} & VaT(Ours) \\ \midrule
     & PSNR $\uparrow$&  26.839  &  26.345    &  27.219           & 27.181             &  \textcolor{red}{27.871}      &    \textcolor{blue}{27.469}    &  26.682           \\
VOCh-test & SSIM $\uparrow$&  0.903   &  0.892          &   0.911          & 0.903           &\textcolor{red}{0.925}    & \textcolor{blue}{0.918}       &  0.901           \\
     & LPIPS $\downarrow$ &  0.069   &   0.084        &    0.058         & 0.061       & \textcolor{red}{0.047}    & \textcolor{blue}{0.053}       & 0.076            \\\midrule
RTTS & NIQE  $\downarrow$& 5.807    &   5.245         &  4.463      &    5.739    &  \textcolor{blue}{4.362}    &  4.376     &  \textcolor{red}{4.191}      \\ \bottomrule
\end{tabular}}
\end{table}

\begin{figure}[t]
    \centering
    \includegraphics[width=0.9\linewidth]{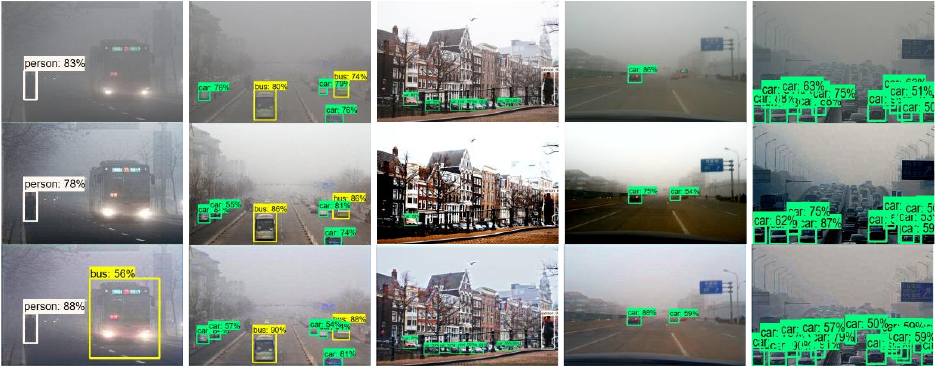}
    \caption{Qualitative comparison of detection results on RTTS \cite{li2018benchmarking}. From top to down are the detection results of the degraded images, the restored images by AONet \cite{li2017aod}, and the translated images by VaT.}
    \label{fig:vision}
\end{figure}

\begin{table}[t]
\centering
\caption{Quantitative comparison results on the ExDark and the VOCd-test. \textcolor{red}{Red} denotes the best results, while \textcolor{blue}{blue} denotes the second-best results.}
\label{tb:exdark_detection}
\resizebox{\linewidth}{!}{
\begin{tabular}{@{}cr|cccc|cccc@{}}
\toprule
                            &      & \multicolumn{4}{c|}{Supervised methods}           & \multicolumn{4}{c}{Unsupervised methods}       \\ 
                            &      & Vanilla (YOLOv5) & Baseline & IA-YOLO \cite{liu2018image} & AD-YOLO \cite{sun2022rethinking} & CycleGAN \cite{zhu2017unpaired} & EnlightenGAN \cite{jiang2021enlightengan} & SRC \cite{jung2022exploring} & VaT (Ours) \\\midrule
\multirow{4}{*}{VOCd} & mAP $\uparrow$  &  0.551  &   0.572  &  0.529  & \textcolor{blue}{0.578}   &  0.446        &0.568      & 0.467    &   \textcolor{red}{0.600 }       \\
                            & PSNR $\uparrow$&  12.018    &  15.478        &  -       &  15.216       &  15.863        & \textcolor{blue}{17.001}      & 16.028    & \textcolor{red}{17.141}          \\
                            & SSIM $\uparrow$&  0.395       &  0.596        &  -       &  0.538       & 0.591         &  \textcolor{blue}{0.630}     &  0.607   & \textcolor{red}{0.633}          \\
                            & LPIPS $\downarrow$&  0.314       & \textcolor{blue}{0.244}         &  -       &  0.271       & 0.259         & 0.244      & 0.268    &  \textcolor{red}{0.215}         \\ \midrule      
\multirow{2}{*}{ExDark}     & mAP  $\uparrow$&  \textcolor{blue}{0.488}  &   0.475  &  0.385  &  0.462  &  0.357        & 0.389    &  0.326   &  \textcolor{red}{0.499}     \\
                            & NIQE $\downarrow$&  5.163       &   4.151   & -   & 4.220        &   4.661       & \textcolor{blue}{4.119}    &  4.447   & \textcolor{red}{3.829}          \\ \bottomrule 
\end{tabular}}
\end{table}

\section{Experiments}
In this section, we conduct experiments on three tasks, including dehazing for detection, low-light enhancement for detection, and low-light enhancement for classification using a large-scale vision-language model.


\subsection{Experimental Setup}
\noindent\textbf{Dehazing Dataset.} RTTS \cite{li2018benchmarking} is a real-world hazy dataset that includes detection annotations for testing purposes. The dataset comprises five object categories: \textit{person, car, bus, bicycle, motorcycle}. We collect a training set, named VOCh, by selecting PASCAL VOC images containing the aforementioned five categories and simulating haze following \cite{sun2022rethinking, liu2022image}, using two scales. Finally, RTTS dataset comprises of 4,322 real-world hazy images. VOCh-train and VOCh-test datasets include 16,222 and 5,468 synthetic hazy images, respectively.

\noindent\textbf{Low-light Enhancement Dataset.}
ExDark \cite{loh2019getting} is a natural low-light dataset with ten object detection classes. We create the training set, VOCd-train, by selecting samples from PASCAL VOC that fall under these categories and simulating low-light corruption \cite{sun2022rethinking} using a gamma value ($\gamma \in [1.5,5]$). ExDark consists of 2,563 real-world dark images, while VOCd-train and VOCd-test include 12,334 and 3,760 synthetic dark images respectively. To address the scarcity of labeled low-light image classification datasets, we generate a training dataset by randomly applying degradation to images from the CIFAR-10 dataset \cite{krizhevsky2009learning}.

\noindent\textbf{Implementation Details.} 
To validate the contribution of our method to the TDIR task, we use AODNet \cite{li2017aod} for dehazing and SCI \cite{ma2022toward} for low-light enhancement as the restoration networks. For detection, we use YOLOv3 \cite{redmon2018yolov3} for dehazing and YOLOv5 for low-light enhancement. These networks are trained on the clean images of VOC-train. For large-scale classification, we use the pre-trained CLIP model \cite{radford2021learning}. No data augmentation is used. All experiments are conducted on a Nvidia RTX4090 24G GPU and implemented in PyTorch. Each task is trained for 50 epochs with a batch size of 4. The learning rate is set to 1e-4, using the AdamW \cite{loshchilov2018decoupled} optimizer with default settings. More details of the experimental settings can be found in the Appendix.



\subsection{Results}
\label{ex:results}
\textbf{Dehazing for Detection.} 
We evaluate our method on dehazing for detection with state-of-the-art supervised TDIR methods \cite{liu2022image, sun2022rethinking} and unsupervised image translation methods \cite{zhu2017unpaired, park2020contrastive, xu2022maximum, jung2022exploring, xie2023unpaired}. The experimental results in Table \ref{tab:COM} show that our approach significantly outperforms unsupervised methods. This is because their methods focus on global transformations of the entire image, while ours focuses on local transformations of target objects, avoiding the effects of non-target objects. VaT achieves comparable results to the supervised methods and even outperforms them on the real-world RTTS dataset. Additionally, our approach is stable whether using pre-trained or retrained AODNet. We further evaluate the visual quality of our approach on image restoration in Table \ref{tb:visual}. On the synthetic VOC-test dataset, VaT and AD-YOLO struggle to enhance image quality, while unpaired methods show marginal improvement. Conversely, on the real-world RTTS dataset, VaT demonstrates the most substantial quality enhancement. Figure \ref{fig:vision} shows some qualitative detection results.

\begin{figure}[t]
    \centering
    \includegraphics[width=0.9\linewidth]{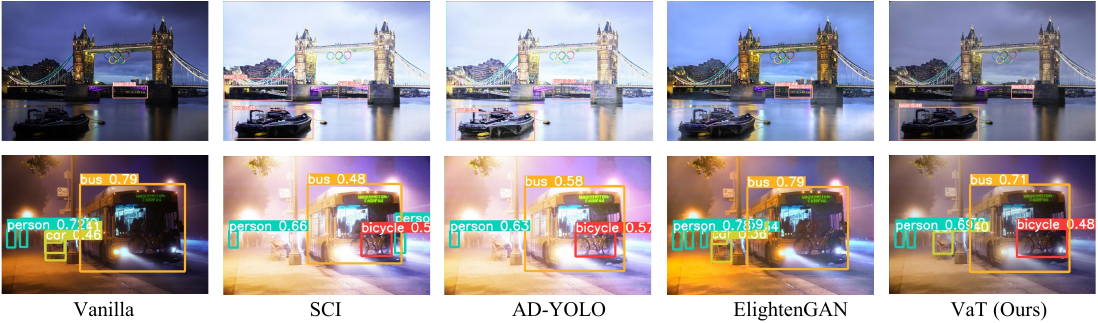}
    \caption{Visual comparison of object detection results on ExDark dataset.}
    \label{fig:vision2}
\end{figure}
\begin{figure}[t]
    \centering
    \includegraphics[width=0.9\linewidth]{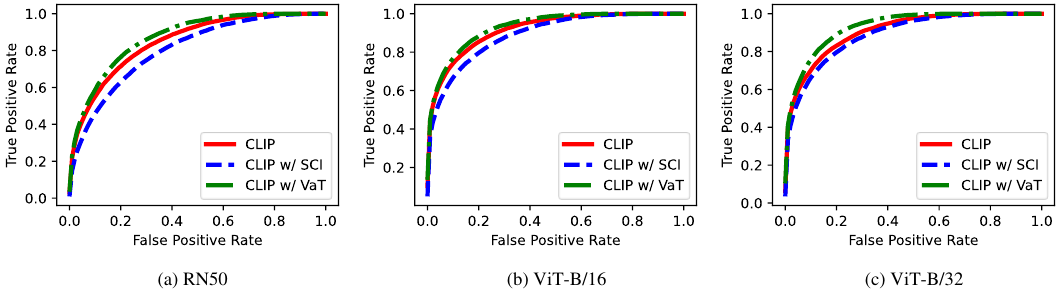}
    \caption{Qualitative comparison results of ROC curve for CLIP \cite{radford2021learning} with different backbones on the low-light CIFAR10 test set.}
    \label{fig:CLIP}
\end{figure}

\noindent\textbf{Low-light Enhancement for Detection.} 
Table \ref{tb:exdark_detection} demonstrates that VaT significantly outperforms other methods in low-light enhancement for detection \cite{jiang2021enlightengan,jung2022exploring,zhu2017unpaired}. Compared to the original YOLOv5, VaT enhances object detection mean average precision (mAP) by 5\% on synthetic datasets and by 3\% compared to the baseline method with SCI for detection. In terms of image restoration, VaT achieves top PSNR and SSIM results on synthetic datasets and shows superior NIQE performance on the ExDark dataset. Object detection results on dark images are visualized in Figure \ref{fig:vision2}, highlighting the superior performance of VaT.

\noindent\textbf{Low-light Enhancement for CLIP Classification.} We chose three sets of pre-trained weights for the CLIP model, and the experimental results are shown in Figure \ref{fig:CLIP}. We can observe that: (1) Directly using low-light enhancement does not effectively improve the performance of the CLIP model, even significantly decreasing the performance. (2) Our method successfully improves the performance of the CLIP model in low-light scenes. (3) Our method has good versatility and improves the original prediction results under three pre-training weights.

\subsection{Ablation Study}
\begin{figure}[t]
    \centering
    \includegraphics[width=0.9\linewidth]{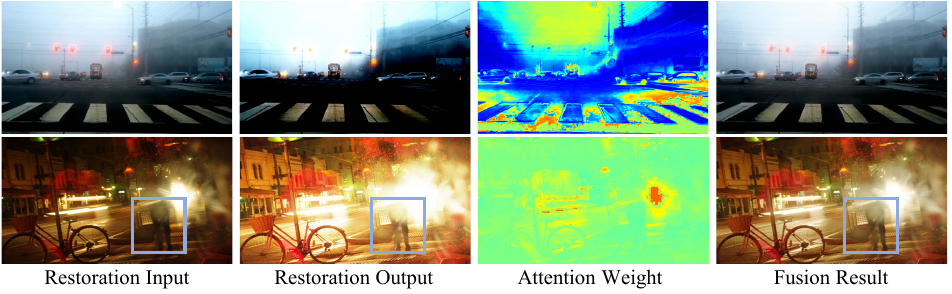}
    \caption{Visualization of gated fusion module process.}
    \label{fig:attention}
\end{figure}

\noindent\textbf{Effectiveness of Core Components.} We validate the effectiveness of each component in VaT, including two loss functions, gated fusion module, Mixup \cite{zhang2018mixup}, and UMix \cite{han2022umix} data augmentation strategy as reported in Table \ref{tab:ablation_}. The first three rows indicate that the absence of consistency loss harms performance, while the absence of the maximum likelihood term loss results in network failure, providing evidence for satisfying Remark \ref{remark:2} in meeting both human and machine visual requirements. The fourth row establishes the significance of the gated fusion module. Additionally, we visualize the outcomes of gated fusion, illustrated in Figure \ref{fig:attention}. The network tends to learn the human-friendly fused outcomes, implicitly demonstrating that superior restoration results can enhance high-level vision performance (Remark \ref{remark:1}). The final two rows indicate that uncertainty-guided mixup \cite{han2022umix} can effectively further enhance performance.

\begin{table}[b]
\centering
\caption{Ablation study of the VaT components. Task1 and Task2 refer to the detection of hazed images (mAP \% $\uparrow$) and the CLIP classification of low-light images (accuracy \% $\uparrow$), respectively.}
\label{tab:ablation_}
\resizebox{0.45\linewidth}{!}{
\begin{tabular}{lllll|ll}
\toprule
\multicolumn{1}{c}{\multirow{2}{*}{\begin{tabular}[c]{@{}c@{}}$\mathcal L_{cycle}$ \end{tabular}}} & \multicolumn{1}{c}{\multirow{2}{*}{\begin{tabular}[c]{@{}c@{}}$\mathcal L_{mle}$ \end{tabular}}} & \multicolumn{1}{c}{\multirow{2}{*}{\begin{tabular}[c]{@{}c@{}}Gated \\ Fusion \end{tabular}}} & \multicolumn{1}{c}{\multirow{2}{*}{\begin{tabular}[c]{@{}c@{}}Mixup \end{tabular}}} & \multicolumn{1}{c|}{\multirow{2}{*}{\begin{tabular}[c]{@{}c@{}}UMix \end{tabular}}} & \multicolumn{1}{c}{\multirow{2}{*}{\begin{tabular}[c]{@{}c@{}}Task1 \end{tabular}}} & \multicolumn{1}{c}{\multirow{2}{*}{\begin{tabular}[c]{@{}c@{}}Task2 \end{tabular}}} \\ 
\multicolumn{1}{c}{}&\multicolumn{1}{c}{}&\multicolumn{1}{c}{}&\multicolumn{1}{c}{}&\multicolumn{1}{c|}{}&\multicolumn{1}{c}{}&\multicolumn{1}{c}{}\\ \midrule
&$\checkmark$&       &       &        &41.97       & 34.85      \\ 
$\checkmark$&       &       &       &        &11.45& 10.42      \\ 
$\checkmark$&$\checkmark$&       &       &        &42.77&35.10       \\ 
$\checkmark$&$\checkmark$&$\checkmark$&       &        &47.49& 41.62      \\ 
$\checkmark$&$\checkmark$&$\checkmark$&$\checkmark$&        &49.12& 43.19\\
$\checkmark$&$\checkmark$&$\checkmark$&&$\checkmark$&\textbf{49.48}&\textbf{44.27}\\ \bottomrule
\end{tabular}}
\end{table}
\begin{figure}[t]
    \centering
    \includegraphics[width=0.85\linewidth]{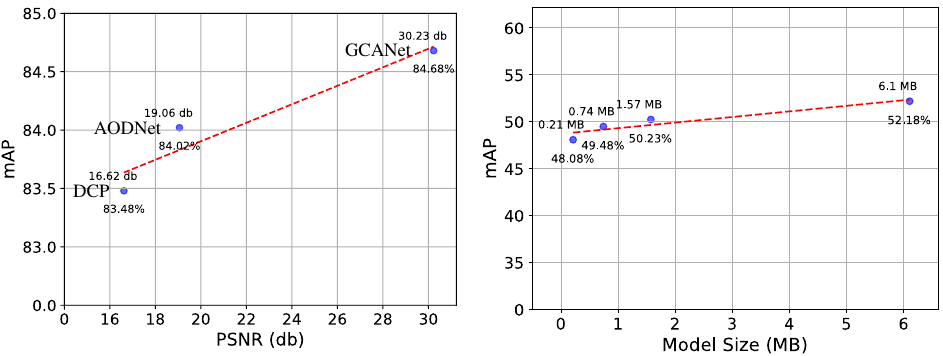}
    \caption{Correlation study. \textit{Left:} Correlation between restoration capability and high-level vision performance on the VOCh-test. \textit{Right:} Correlation between translator size and high-level vision performance on the RTTS.}
    \label{fig:relation}
\end{figure}
\noindent\textbf{Effectiveness of Restoration Capability.} To verify whether our method complies with Remark \ref{remark:1}, we select different restoration algorithms to evaluate the impact on the high-level vision performance. Specifically, we select DCP \cite{he2010single}, AODNet \cite{li2017aod}, and GCANet \cite{chen2019gated} as the restoration algorithms and use PSNR in their paper as the evaluation metric for image restoration performance. The experimental results are shown in the left of Figure \ref{fig:relation}, indicating a positive correlation between image restoration performance and high-level vision performance. Therefore, it validates that our method can effectively leverage the performance of the restoration model.

\noindent\textbf{Effectiveness of Translator Capability.} We set different base dimensions for the transformer block \cite{zamir2022restormer} to simulate different fitting abilities for VaT. Specifically, we chose five cases [4, 8, 16, 32, 48], corresponding to model sizes of [0.21M, 0.74M, 1.57M, 2.75M, 6.10M]. The experimental results are shown in the right of Figure \ref{fig:relation}, we can observe that there is a positive correlation between translator model size and performance. Therefore, improving translator models of VaT is a direction worth exploring in the future but it is not the focus of this paper.




\section{Conclusion}
In this paper, we aim to bridge the gap between image restoration and high-level vision tasks by establishing the joint distribution between them. This is achieved through the simplification of the optimization objective using variational and full probability formulas. To validate our objective, we propose an unsupervised method called VaT, which includes an efficient translator network and utilizes cycle-consistency training and self-training. Our method undergoes extensive evaluation on three paired restoration and high-level vision tasks, demonstrating superior performance compared to counterpart methods. In the future, we will further expand VaT to multimodal large language models.

\clearpage  

\section*{Acknowledgements}
This work was supported by the National Natural Science Foundation of China under Grant No. 62071500; Supported by Supported by Shenzhen Science and Technology Program under Grant No. JCYJ20230807111107015. Supported by Fundamental Research Funds for the Central Universities, Sun Yat-sen University under Grant No. 241gqb015.

%
%
\bibliographystyle{splncs04}
\bibliography{main}

\begin{thebibliography}{10}
\providecommand{\url}[1]{\texttt{#1}}
\providecommand{\urlprefix}{URL }
\providecommand{\doi}[1]{https://doi.org/#1}

\bibitem{banham1997digital}
Banham, M.R., Katsaggelos, A.K.: Digital image restoration. IEEE signal processing magazine  \textbf{14}(2),  24--41 (1997)

\bibitem{benaim2017one}
Benaim, S., Wolf, L.: One-sided unsupervised domain mapping. Advances in neural information processing systems  \textbf{30} (2017)

\bibitem{berman2016non}
Berman, D., Avidan, S., et~al.: Non-local image dehazing. In: Proceedings of the IEEE conference on computer vision and pattern recognition. pp. 1674--1682 (2016)

\bibitem{blei2017variational}
Blei, D.M., Kucukelbir, A., McAuliffe, J.D.: Variational inference: A review for statisticians. Journal of the American statistical Association  \textbf{112}(518),  859--877 (2017)

\bibitem{chen2019gated}
Chen, D., He, M., Fan, Q., Liao, J., Zhang, L., Hou, D., Yuan, L., Hua, G.: Gated context aggregation network for image dehazing and deraining. In: 2019 IEEE winter conference on applications of computer vision (WACV). pp. 1375--1383. IEEE (2019)

\bibitem{chen2022unpaired}
Chen, X., Fan, Z., Li, P., Dai, L., Kong, C., Zheng, Z., Huang, Y., Li, Y.: Unpaired deep image dehazing using contrastive disentanglement learning. In: European Conference on Computer Vision. pp. 632--648. Springer (2022)

\bibitem{chen2022unpaired1}
Chen, X., Pan, J., Jiang, K., Li, Y., Huang, Y., Kong, C., Dai, L., Fan, Z.: Unpaired deep image deraining using dual contrastive learning. In: Proceedings of the IEEE/CVF Conference on Computer Vision and Pattern Recognition. pp. 2017--2026 (2022)

\bibitem{chen2022vector}
Chen, Y.J., Cheng, S.I., Chiu, W.C., Tseng, H.Y., Lee, H.Y.: Vector quantized image-to-image translation. In: European Conference on Computer Vision. pp. 440--456. Springer (2022)

\bibitem{dai2016image}
Dai, D., Wang, Y., Chen, Y., Van~Gool, L.: Is image super-resolution helpful for other vision tasks? In: 2016 IEEE Winter Conference on Applications of Computer Vision (WACV). pp.~1--9. IEEE (2016)

\bibitem{dalva2022vecgan}
Dalva, Y., Alt{\i}ndi{\c{s}}, S.F., Dundar, A.: Vecgan: Image-to-image translation with interpretable latent directions. In: European Conference on Computer Vision. pp. 153--169. Springer (2022)

\bibitem{fu2019geometry}
Fu, H., Gong, M., Wang, C., Batmanghelich, K., Zhang, K., Tao, D.: Geometry-consistent generative adversarial networks for one-sided unsupervised domain mapping. In: Proceedings of the IEEE/CVF Conference on Computer Vision and Pattern Recognition. pp. 2427--2436 (2019)

\bibitem{guo2023low}
Guo, X., Hu, Q.: Low-light image enhancement via breaking down the darkness. International Journal of Computer Vision  \textbf{131}(1),  48--66 (2023)

\bibitem{han2021dual}
Han, J., Shoeiby, M., Petersson, L., Armin, M.A.: Dual contrastive learning for unsupervised image-to-image translation. In: Proceedings of the IEEE/CVF conference on computer vision and pattern recognition. pp. 746--755 (2021)

\bibitem{han2022umix}
Han, Z., Liang, Z., Yang, F., Liu, L., Li, L., Bian, Y., Zhao, P., Wu, B., Zhang, C., Yao, J.: Umix: Improving importance weighting for subpopulation shift via uncertainty-aware mixup. Advances in Neural Information Processing Systems  \textbf{35},  37704--37718 (2022)

\bibitem{he2010single}
He, K., Sun, J., Tang, X.: Single image haze removal using dark channel prior. IEEE transactions on pattern analysis and machine intelligence  \textbf{33}(12),  2341--2353 (2010)

\bibitem{huang2018multimodal}
Huang, X., Liu, M.Y., Belongie, S., Kautz, J.: Multimodal unsupervised image-to-image translation. In: Proceedings of the European conference on computer vision (ECCV). pp. 172--189 (2018)

\bibitem{jia2021semantically}
Jia, Z., Yuan, B., Wang, K., Wu, H., Clifford, D., Yuan, Z., Su, H.: Semantically robust unpaired image translation for data with unmatched semantics statistics. In: Proceedings of the IEEE/CVF International Conference on Computer Vision. pp. 14273--14283 (2021)

\bibitem{jiang2021enlightengan}
Jiang, Y., Gong, X., Liu, D., Cheng, Y., Fang, C., Shen, X., Yang, J., Zhou, P., Wang, Z.: Enlightengan: Deep light enhancement without paired supervision. IEEE transactions on image processing  \textbf{30},  2340--2349 (2021)

\bibitem{jung2022exploring}
Jung, C., Kwon, G., Ye, J.C.: Exploring patch-wise semantic relation for contrastive learning in image-to-image translation tasks. In: Proceedings of the IEEE/CVF conference on computer vision and pattern recognition. pp. 18260--18269 (2022)

\bibitem{kingma2014auto}
Kingma, D.P., Welling, M.: Auto-encoding variational bayes. stat  \textbf{1050}, ~1 (2014)

\bibitem{kotelevskii2022nonparametric}
Kotelevskii, N., Artemenkov, A., Fedyanin, K., Noskov, F., Fishkov, A., Shelmanov, A., Vazhentsev, A., Petiushko, A., Panov, M.: Nonparametric uncertainty quantification for single deterministic neural network. Advances in Neural Information Processing Systems  \textbf{35},  36308--36323 (2022)

\bibitem{krizhevsky2009learning}
Krizhevsky, A., Hinton, G., et~al.: Learning multiple layers of features from tiny images  (2009)

\bibitem{kullback1951information}
Kullback, S., Leibler, R.A.: On information and sufficiency. The annals of mathematical statistics  \textbf{22}(1),  79--86 (1951)

\bibitem{lee2018diverse}
Lee, H.Y., Tseng, H.Y., Huang, J.B., Singh, M., Yang, M.H.: Diverse image-to-image translation via disentangled representations. In: Proceedings of the European conference on computer vision (ECCV). pp. 35--51 (2018)

\bibitem{li2017aod}
Li, B., Peng, X., Wang, Z., Xu, J., Feng, D.: Aod-net: All-in-one dehazing network. In: Proceedings of the IEEE international conference on computer vision. pp. 4770--4778 (2017)

\bibitem{li2018benchmarking}
Li, B., Ren, W., Fu, D., Tao, D., Feng, D., Zeng, W., Wang, Z.: Benchmarking single-image dehazing and beyond. IEEE Transactions on Image Processing  \textbf{28}(1),  492--505 (2018)

\bibitem{li2023detection}
Li, C., Zhou, H., Liu, Y., Yang, C., Xie, Y., Li, Z., Zhu, L.: Detection-friendly dehazing: Object detection in real-world hazy scenes. IEEE Transactions on Pattern Analysis and Machine Intelligence  (2023)

\bibitem{liu2018unified}
Liu, A.H., Liu, Y.C., Yeh, Y.Y., Wang, Y.C.F.: A unified feature disentangler for multi-domain image translation and manipulation. Advances in neural information processing systems  \textbf{31} (2018)

\bibitem{liu2020connecting}
Liu, D., Wen, B., Jiao, J., Liu, X., Wang, Z., Huang, T.S.: Connecting image denoising and high-level vision tasks via deep learning. IEEE Transactions on Image Processing  \textbf{29},  3695--3706 (2020)

\bibitem{liu2018image}
Liu, D., Wen, B., Liu, X., Wang, Z., Huang, T.S.: When image denoising meets high-level vision tasks: a deep learning approach. In: Proceedings of the 27th International Joint Conference on Artificial Intelligence. pp. 842--848 (2018)

\bibitem{liu2022image}
Liu, W., Ren, G., Yu, R., Guo, S., Zhu, J., Zhang, L.: Image-adaptive yolo for object detection in adverse weather conditions. In: Proceedings of the AAAI Conference on Artificial Intelligence. vol.~36, pp. 1792--1800 (2022)

\bibitem{liu2023degae}
Liu, Y., He, J., Gu, J., Kong, X., Qiao, Y., Dong, C.: Degae: A new pretraining paradigm for low-level vision. In: Proceedings of the IEEE/CVF Conference on Computer Vision and Pattern Recognition. pp. 23292--23303 (2023)

\bibitem{liu2023evaluating}
Liu, Y., Zhao, H., Gu, J., Qiao, Y., Dong, C.: Evaluating the generalization ability of super-resolution networks. IEEE Transactions on pattern analysis and machine intelligence  (2023)

\bibitem{loh2019getting}
Loh, Y.P., Chan, C.S.: Getting to know low-light images with the exclusively dark dataset. Computer Vision and Image Understanding  \textbf{178},  30--42 (2019)

\bibitem{loshchilov2018decoupled}
Loshchilov, I., Hutter, F.: Decoupled weight decay regularization. In: International Conference on Learning Representations (2018)

\bibitem{ma2022toward}
Ma, L., Ma, T., Liu, R., Fan, X., Luo, Z.: Toward fast, flexible, and robust low-light image enhancement. In: Proceedings of the IEEE/CVF Conference on Computer Vision and Pattern Recognition. pp. 5637--5646 (2022)

\bibitem{ozdenizci2023restoring}
{\"O}zdenizci, O., Legenstein, R.: Restoring vision in adverse weather conditions with patch-based denoising diffusion models. IEEE Transactions on Pattern Analysis and Machine Intelligence  (2023)

\bibitem{pang2021image}
Pang, Y., Lin, J., Qin, T., Chen, Z.: Image-to-image translation: Methods and applications. IEEE Transactions on Multimedia  \textbf{24},  3859--3881 (2021)

\bibitem{park2020contrastive}
Park, T., Efros, A.A., Zhang, R., Zhu, J.Y.: Contrastive learning for unpaired image-to-image translation. In: Computer Vision--ECCV 2020: 16th European Conference, Glasgow, UK, August 23--28, 2020, Proceedings, Part IX 16. pp. 319--345. Springer (2020)

\bibitem{pei2018does}
Pei, Y., Huang, Y., Zou, Q., Lu, Y., Wang, S.: Does haze removal help cnn-based image classification? In: Proceedings of the European Conference on Computer Vision (ECCV). pp. 682--697 (2018)

\bibitem{radford2021learning}
Radford, A., Kim, J.W., Hallacy, C., Ramesh, A., Goh, G., Agarwal, S., Sastry, G., Askell, A., Mishkin, P., Clark, J., et~al.: Learning transferable visual models from natural language supervision. In: International conference on machine learning. pp. 8748--8763. PMLR (2021)

\bibitem{redmon2018yolov3}
Redmon, J.: Yolov3: An incremental improvement. CoRR  (2018)

\bibitem{ronneberger2015u}
Ronneberger, O., Fischer, P., Brox, T.: U-net: Convolutional networks for biomedical image segmentation. In: Medical Image Computing and Computer-Assisted Intervention--MICCAI 2015: 18th International Conference, Munich, Germany, October 5-9, 2015, Proceedings, Part III 18. pp. 234--241. Springer (2015)

\bibitem{singh2019dual}
Singh, M., Nagpal, S., Singh, R., Vatsa, M.: Dual directed capsule network for very low resolution image recognition. In: Proceedings of the IEEE/CVF International Conference on Computer Vision. pp. 340--349 (2019)

\bibitem{sohn2020fixmatch}
Sohn, K., Berthelot, D., Carlini, N., Zhang, Z., Zhang, H., Raffel, C.A., Cubuk, E.D., Kurakin, A., Li, C.L.: Fixmatch: Simplifying semi-supervised learning with consistency and confidence. Advances in neural information processing systems  \textbf{33},  596--608 (2020)

\bibitem{song2023vision}
Song, Y., He, Z., Qian, H., Du, X.: Vision transformers for single image dehazing. IEEE Transactions on Image Processing  \textbf{32},  1927--1941 (2023)

\bibitem{sun2022rethinking}
Sun, S., Ren, W., Wang, T., Cao, X.: Rethinking image restoration for object detection. Advances in Neural Information Processing Systems  \textbf{35},  4461--4474 (2022)

\bibitem{tang2021attentiongan}
Tang, H., Liu, H., Xu, D., Torr, P.H., Sebe, N.: Attentiongan: Unpaired image-to-image translation using attention-guided generative adversarial networks. IEEE transactions on neural networks and learning systems  (2021)

\bibitem{vaswani2017attention}
Vaswani, A., Shazeer, N., Parmar, N., Uszkoreit, J., Jones, L., Gomez, A.N., Kaiser, {\L}., Polosukhin, I.: Attention is all you need. Advances in neural information processing systems  \textbf{30} (2017)

\bibitem{wu2022uretinex}
Wu, W., Weng, J., Zhang, P., Wang, X., Yang, W., Jiang, J.: Uretinex-net: Retinex-based deep unfolding network for low-light image enhancement. In: Proceedings of the IEEE/CVF conference on computer vision and pattern recognition. pp. 5901--5910 (2022)

\bibitem{xie2021unaligned}
Xie, S., Gong, M., Xu, Y., Zhang, K.: Unaligned image-to-image translation by learning to reweight. In: Proceedings of the IEEE/CVF International Conference on Computer Vision. pp. 14174--14184 (2021)

\bibitem{xie2022unsupervised}
Xie, S., Ho, Q., Zhang, K.: Unsupervised image-to-image translation with density changing regularization. Advances in Neural Information Processing Systems  \textbf{35},  28545--28558 (2022)

\bibitem{xie2023unpaired}
Xie, S., Xu, Y., Gong, M., Zhang, K.: Unpaired image-to-image translation with shortest path regularization. In: Proceedings of the IEEE/CVF Conference on Computer Vision and Pattern Recognition. pp. 10177--10187 (2023)

\bibitem{xu2022maximum}
Xu, Y., Xie, S., Wu, W., Zhang, K., Gong, M., Batmanghelich, K.: Maximum spatial perturbation consistency for unpaired image-to-image translation. In: Proceedings of the IEEE/CVF Conference on Computer Vision and Pattern Recognition. pp. 18311--18320 (2022)

\bibitem{yang2022self}
Yang, Y., Wang, C., Liu, R., Zhang, L., Guo, X., Tao, D.: Self-augmented unpaired image dehazing via density and depth decomposition. In: Proceedings of the IEEE/CVF conference on computer vision and pattern recognition. pp. 2037--2046 (2022)

\bibitem{yosida2012functional}
Yosida, K.: Functional analysis. Springer Science \& Business Media (2012)

\bibitem{zamir2022restormer}
Zamir, S.W., Arora, A., Khan, S., Hayat, M., Khan, F.S., Yang, M.H.: Restormer: Efficient transformer for high-resolution image restoration. In: Proceedings of the IEEE/CVF conference on computer vision and pattern recognition. pp. 5728--5739 (2022)

\bibitem{zhang2018mixup}
Zhang, H., Cisse, M., Dauphin, Y.N., Lopez-Paz, D.: mixup: Beyond empirical risk minimization. In: International Conference on Learning Representations (2018)

\bibitem{zhu2017unpaired}
Zhu, J.Y., Park, T., Isola, P., Efros, A.A.: Unpaired image-to-image translation using cycle-consistent adversarial networks. In: Proceedings of the IEEE international conference on computer vision. pp. 2223--2232 (2017)

\end{thebibliography}
\end{document}